\documentclass[letterpaper, 10 pt, conference]{ieeeconf}
\IEEEoverridecommandlockouts
\usepackage{cite}
\usepackage{amsmath,amssymb,amsfonts}
\usepackage{algorithmic}
\usepackage{graphicx}
\usepackage{textcomp}
\usepackage{xcolor}
\usepackage{url}
\usepackage{placeins}
\usepackage{float}
\usepackage{tabularx,colortbl}
\usepackage{ifthen}
\usepackage{caption,subcaption}
\usepackage{amsmath}
\usepackage{booktabs}
\renewcommand{\thetable}{\Roman{table}}
\usepackage{cleveref}
\let\labelindent\relax
\usepackage{enumitem}
\usepackage{stfloats}
\usepackage{wrapfig}
\usepackage{verbatim}
\def\BibTeX{{\rm B\kern-.05em{\sc i\kern-.025em b}\kern-.08em
    T\kern-.1667em\lower.7ex\hbox{E}\kern-.125emX}}

\begin{document}
\newcolumntype{P}[1]{>{\centering\arraybackslash}p{#1}}
\newcolumntype{M}[1]{>{\centering\arraybackslash}m{#1}}
\setlength{\textfloatsep}{10pt plus 1.0pt minus 3.0pt}
\setlength{\dbltextfloatsep}{10pt plus 1.0pt minus 3.0pt}
\setlength{\floatsep}{10pt plus 1.0pt minus 3.0pt}
\setlength{\dblfloatsep}{10pt plus 1.0pt minus 3.0pt}
\setlength{\intextsep}{10pt plus 1.0pt minus 3.0pt}
\title{\LARGE \bf Lightweight, Uncertainty-Aware Conformalized Visual Odometry}
\author{Alex C. Stutts, Danilo Erricolo, Theja Tulabandhula, and Amit Ranjan Trivedi
\thanks{Authors are with the University of Illinois Chicago (UIC), Chicago, IL, Email: {\tt\small astutt2@uic.edu}}
}
\maketitle

\begin{abstract} 
Data-driven visual odometry (VO) is a critical subroutine for autonomous edge robotics, and recent progress in the field has produced highly accurate point predictions in complex environments. However, emerging autonomous edge robotics devices like insect-scale drones and surgical robots lack a computationally efficient framework to estimate VO's predictive uncertainties. Meanwhile, as edge robotics continue to proliferate into mission-critical application spaces, awareness of model's the predictive uncertainties has become crucial for risk-aware decision-making. This paper addresses this challenge by presenting a novel, lightweight, and statistically robust framework that leverages conformal inference (CI) to extract VO's uncertainty bands. Our approach represents the uncertainties using flexible, adaptable, and adjustable prediction intervals that, on average, guarantee the inclusion of the ground truth across all degrees of freedom (DOF) of pose estimation. We discuss the architectures of generative deep neural networks for estimating multivariate uncertainty bands along with point (mean) prediction. We also present techniques to improve the uncertainty estimation accuracy, such as leveraging Monte Carlo dropout (MC-dropout) for data augmentation. Finally, we propose a novel training loss function that combines interval scoring and calibration loss with traditional training metrics--mean-squared error and KL-divergence--to improve uncertainty-aware learning. Our simulation results demonstrate that the presented framework consistently captures true uncertainty in pose estimations across different datasets, estimation models, and applied noise types, indicating its wide applicability.
\end{abstract}

\section{Introduction}
As machine learning (ML) becomes more prevalent in privacy, safety, and mission-critical robotics, the ability to quantitatively and visually assess predictive uncertainties of ML models is becoming essential for risk-aware control and decision-making. Two types of uncertainty can result from data-driven learning: \textit{epistemic} and \textit{aleatoric}. Epistemic uncertainty arises from the variance in the training data and can often be reduced with more data. On the other hand, aleatoric uncertainty arises from random distortions in the data, such as blurriness, occlusions, overexposure, \textit{etc.}, which additional training data cannot mitigate.

Although mitigating epistemic uncertainty is challenging, detecting, explaining, and handling aleatoric uncertainty is even more difficult. Therefore, for mission-critical risk-aware robotics, it is desired that predictive models under aleatoric uncertainty must provide confidence measures and express input/model-dependent predictive uncertainties, along with the point (mean) prediction. Additionally, the uncertainty estimates must be extracted with minimal additional computing cost since many edge robotic systems have limited computing and storage capacity due to cost, footprint, legacy design, battery power, and other factors.

Meanwhile, given their computational intensity, traditional methods for uncertainty-aware predictions, such as Bayesian ML models, are unsuited for edge robotics. For example, in Bayesian ML models, a posterior distribution of weights is learned from the training data using Bayesian principles. Weights are then sampled from the posterior, and the network output is statistically computed and weighed against each sample's probability. Generating uncertainty-aware predictions, thus, involves sampling many model parameters from the posterior and evaluating predictions at each sample, which can be impractical under typical time and resource constraints for various edge robotics applications.

\begin{figure}[!t]
\centering
    \includegraphics[width=\columnwidth]{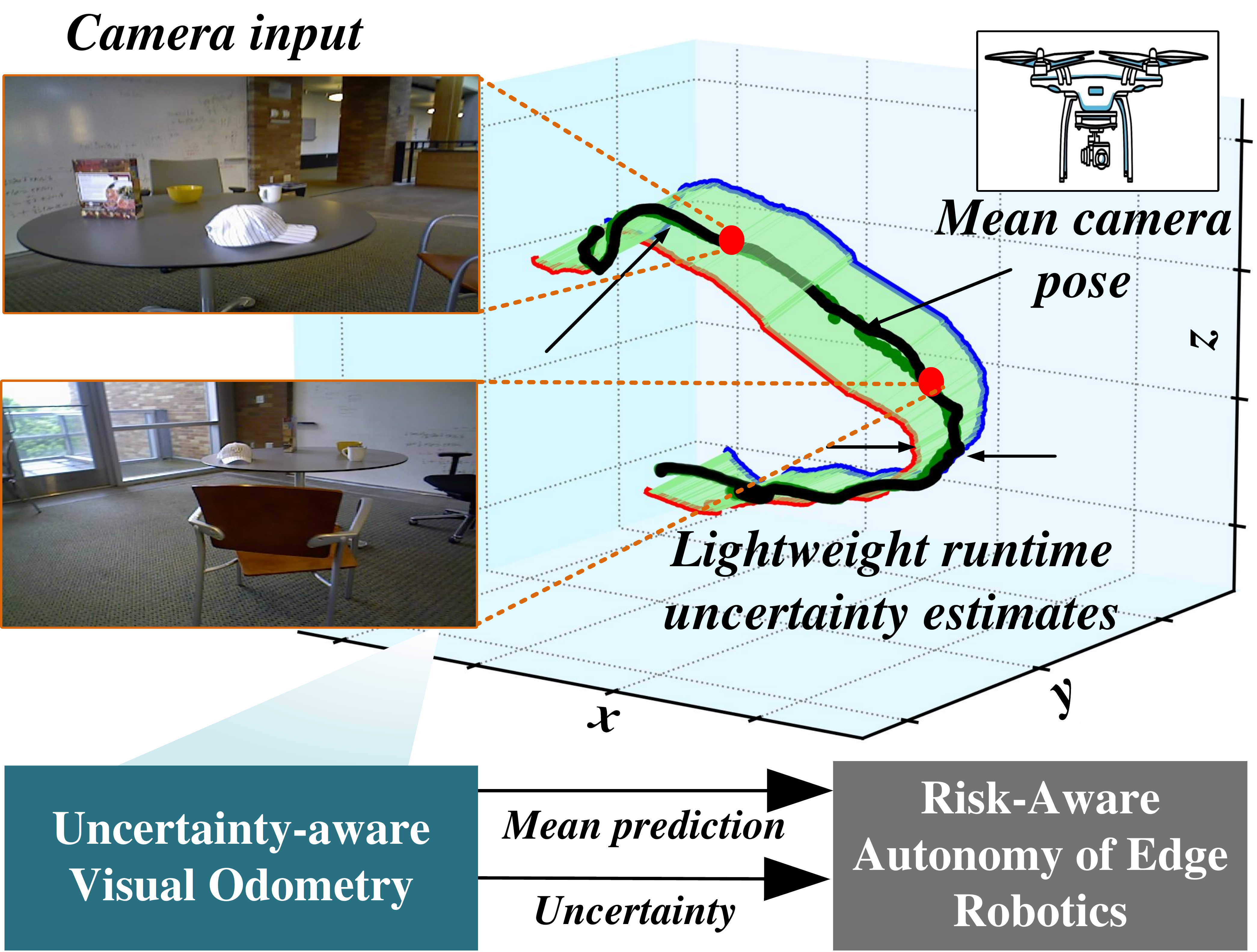}
    \caption{\textbf{Lightweight uncertainty-aware visual odometry (VO):} In this paper, we present a lightweight framework for uncertainty-aware visual odometry for edge robotics where computing resources are limited while the predictions need to be made in real time. Our framework exploits \textit{conformal inference} and presents four novel methodologies to extract point prediction and upper/lower bounds under the designated uncertainty coverage.} 
    \label{fig:titlepage}
\end{figure}

A more computationally-efficient alternative to Bayesian inference is Variational inference (VI) \cite{blei2017}. VI considers a family of approximate densities $F$, from which a member $q^{(w)}$ is learned by minimizing the Kullback-Leibler (KL) divergence to the posterior. However, selecting a flexible family that closely matches the true posterior in practice can be challenging. Additionally, VI struggles to approximate complex posteriors with multiple modes or sharp peaks. 

Recently, conformal inference (CI) of ML models, also known as conformal prediction, was developed to overcome the above limitations of traditional uncertainty-aware prediction frameworks\cite{vovk2005,lei2016,shafer2007,angelopoulos2021,einbinder2022,messoudi2020}. Unlike classical statistical inference, which relies on data distribution to capture prediction uncertainty and can be sensitive to model misspecification, CI is a distribution-free approach that guarantees the statistical validity of uncertainty intervals given a finite training sample. CI can also assess the degree of conformity of each new observation to the available data and uses this information to construct an uncertainty interval calibrated to a user-specified coverage rate. Moreover, CI can be combined with any underlying model with an inherent notion of uncertainty to provide uncertainty quantification that is both statistically valid and model-agnostic. One of the challenges in making CI practical is that the uncertainty estimates are often conservative, and our work overcomes this and other limitations, providing a compelling case for their use in robotics applications with stringent resource usage requirements.

In particular, leveraging these advantages of CI for edge robotics, in this work, we investigate the conformalization of visual odometry (VO), an essential task for autonomous navigation that estimates the position and orientation (i.e., pose) of a camera mounted on a moving vehicle relative to the environment as the camera moves. The resulting motion estimates from VO are used for typical autonomy objectives such as three-dimensional (3D) reconstruction, mapping, localization, \textit{etc.} Especially since VO relies solely on cameras, which are passive, low power, and have a small footprint, VO-based motion estimates are suitable for ego-motion tracking under stringent footprint and battery power constraints. Fig.~\ref{fig:titlepage} overviews the proposed techniques.

Exploring CI for lightweight uncertainty estimates in VO, our work makes the following key contributions: 
\begin{itemize}[noitemsep,topsep=0pt,leftmargin=4pt]
\item We present \textit{four frameworks} for extracting VO-uncertainty bands by univariate and multivariate conformalization. The presented frameworks vary in computing workload, uncertainty accuracy, interval adaptiveness, data dependency, \textit{etc.}, aiming to provide an entire spectrum of solutions for extracting VO's uncertainty estimates under varying computing resources, training data, and timing constraints for edge robotics.      
\item We present a novel loss function for joint training of point (mean) prediction and heteroskedastic uncertainty bounds. The loss function combines an interval score function and combined calibration loss function with mean squared error (MSE) loss and Kullback-Leibler (KL) divergence. Our novel loss function improves the accuracy of uncertainty coverage and enables computationally efficient predictive architecture for the simultaneous extraction of point predictions and uncertainties.    
\item We also discuss data augmentation for uncertainty-aware VO by providing additional training set by Monte Carlo Dropout (MC-Dropout), thus enabling the distillation of uncertainty estimates from sophisticated, computationally expensive frameworks to conformal quantile bounds.
\end{itemize}

\section{Related Works}
Under VO, the camera's ego motion is estimated based on the changes in visual information captured by the camera. The traditional methods utilized techniques that involved identifying and tracking distinctive visual features, such as corners or edges, for VO \cite{scaramuzza2011visual}. Direct methods \cite{alismail2016direct} estimated the camera's motion by analyzing the changes in the intensity values of the pixels in the images captured by the camera. Hybrid methods for VO combined feature-based and direct methods; for example, feature-based techniques were used for initial camera motion and then refined using direct methods \cite{krombach2017combining}. Structure from motion (SfM) is popularly used for simultaneously estimating the 3D structure of the environment and camera's motion \cite{grater2015robust}. 

A recent trend for VO is to use deep neural networks (DNN) to directly learn the relation between the camera's visual field and its ego-motion from data. PoseNet \cite{kendall2015} demonstrated the feasibility of training convolutional neural networks for end-to-end tracking of the ego-motion of monocular cameras. PoseLSTM \cite{walch2017} utilized long-short-term memories (LSTM) to improve ego-motion accuracy compared to frame-based methods such as PoseNet. DeepVO \cite{wang2017} demonstrated the combined strength of recurrent convolutional neural networks (RCNNs) in learning features and modeling sequential relationships between consecutive camera frames. UnDeepVO \cite{li2018} proposed an unsupervised deep learning approach capable of absolute scale recovery. 

While many prior studies have focused on improving the accuracy of VO, only a few have addressed the challenge of extracting its predictive uncertainties, especially under time and computing resource constraints; for instance, integrated deep learning-based depth predictions with particle filtering to achieve uncertainty-aware visual localization \cite{shukla2022robust}. However, particle filtering-based uncertainty quantification can be computationally expensive for large or high-dimensional state spaces. Kalman filters are a more computationally efficient alternative to particle filters, but their usage is limited to linear systems with Gaussian noise assumptions. By integrating Monte-Carlo Dropout (MC-dropout) \cite{gal2015} with deep learning-based predictors, such as PoseNet, \cite{costante2020} showed uncertainty-aware VO estimations. However, the method requires performing a sufficient number of dropout iterations and evaluating predictions for each, making it challenging to implement under time and computing resource constraints. Besides, MC-dropout, as a basis for uncertainty, lacks statistical rigor and ultimately leads to conservative prediction intervals that are not particularly adaptable to the data. D3VO \cite{yang2020} also utilized Bayesian techniques for uncertainty quantification, demonstrating state-of-the-art accuracy but with a large computational expense. With similar computational overhead, MDN-VO \cite{kaygusuz2021} combined an RCNN and mixture density network for uncertainty estimations based on maximizing the likelihood of pose estimations across a sequence of images. In contrast, the authors in \cite{lee2022} presented a lightweight deterministic uncertainty estimator as a small neural network that could be applied to VO or any data-driven deep neural network model. The method uncovers spatial and semantic model uncertainty with significantly less computation but lacks statistical guarantees.

Unlike the above, our methods build on conformal inference, which can provide statistically valid uncertainty intervals while not requiring heavy computational budgets. Combining CI with an underlying model with an inherent notion of uncertainty (e.g., conditional quantile regression) guarantees true value coverage within the uncertainty intervals even when the model's predictions are poor. However, CI alone is not highly adaptable to heteroskedastic data, as prediction sets can appear fixed and weakly dependent on the model's predictors. For example, conformal classification can produce rather conservative prediction sets given their sole reliance on softmax scores, as mentioned in \cite{angelopoulos2021}. For regression problems, conformalized quantile regression (CQR) was introduced in \cite{romano2019}. CQR produces adaptive and distribution-free prediction intervals with a guaranteed marginal coverage rate, such as 90\%. While typical full conformal prediction assumes that samples are drawn exchangeably (i.e., conditionally, i.i.d, on the joint probability distribution function between input and labels), CQR departs from this assumption to achieve a finite sample coverage guarantee through a technique called split conformal prediction. Various forms of CI, including conformal classification and CQR, were initially developed for one-dimensional (1D) data. They have since been extended to multivariate cases, making them particularly valuable for our application to VO.

\begin{table*}[t]
\setlength{\tabcolsep}{1pt}
\centering
\small
\begin{tabular}{*{5}{M{35mm}}}
\multicolumn{5}{c}{\textbf{Table I: Qualitative comparison of proposed techniques on VO-uncertainty band extraction}} \vspace{5pt}\\
\hline
\textbf{Method} & \textbf{CQR} (Sec. III-A) & \textbf{CSP} (Sec. III-B) & \textbf{MCQR w/ MC-dropout} (Sec. IV-A) & \textbf{CJP} (Sec. IV-B) \\ 
\hline
Uncertainty Bands & Univariate & Univariate & Multivariate & Multivariate \\
\hline
Disjoint Bands & No & Yes & No & No \\
\hline
Computational Complexity & Low & Low & High & Medium \\
\hline
Interval Calibration & Fixed & Fixed & Fixed & Tunable \\
\hline
Interval Adaptiveness & Conservative & Semi-tunable & Flexible & Tunable \\
\hline
Uncertainty Accuracy & Low & Medium & High & High \\
\hline
Training Time & Short & Medium & Long & Medium \\
\hline
Data Dependency & Low & High & High & Low \\
\hline
\end{tabular}
\end{table*}

\section{Extracting Univariate Bands of Uncertainties for Visual Odometry (VO)}
This section introduces two methods for extracting predictive uncertainty bands in VO: univariate conformalized quantile regression and conformalized set prediction. The methods presented here are lightweight but produce rectangular uncertainty bands, which can be too conservative. The first method uses quantile regression on each coordinate of the position and orientation output vectors of VO, which are combined to produce an overall quantile region. The second method maps the regression problem in VO to a classification problem and performs a set prediction, which can produce disjoint uncertainty bands. The presented techniques are compared in Table I and discussed below:

\subsection{VO-Uncertainty by Univariate CQR}
\begin{figure}[!t]
\centering
    \includegraphics[width=\columnwidth]{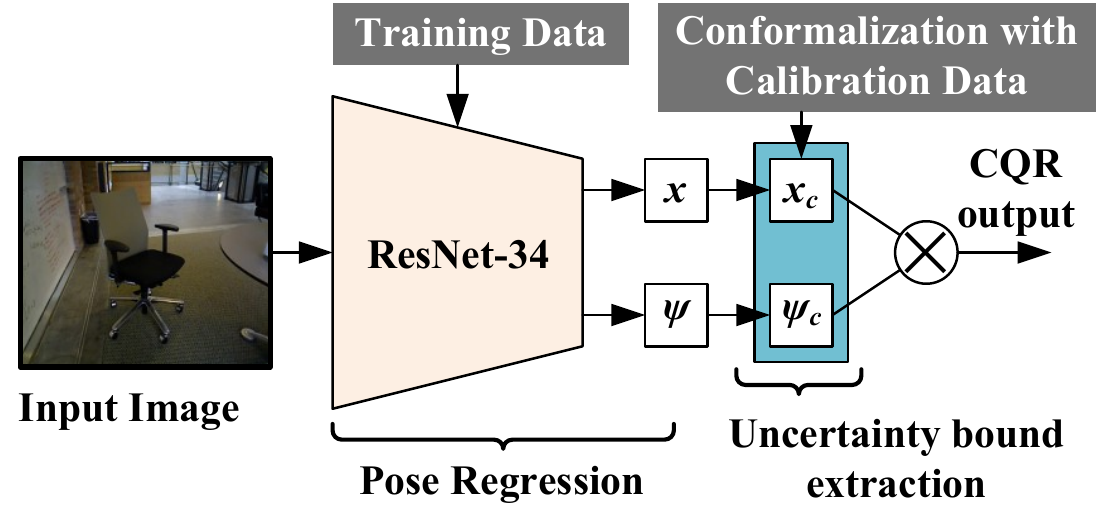}
    \caption{\textbf{VO uncertainty estimation using univariate CQR:} Features of the input image and extracted and regressed to six-dimensional pose. Subsequently, each pose dimension's predictions are conformalized using the held-out calibration data. Finally, multivariate uncertainty regions are predicted by multiplying 1D uncertainty bands.}
    \label{fig:CQR}
\end{figure}

We employ conformalized quantile regression (CQR) \cite{romano2019} to extract univariate uncertainty bands for each output coordinate in VO. To test our approach, we use PoseNet \cite{kendall2015} with a ResNet34-based feature extractor for conformalization. However, the presented techniques are generalizable to other feature-based and direct predictors for VO. The PoseNet model outputs translational elements $x$, $y$, and $z$, as well as rotational elements $p$, $q$, and $r$ of angle $w$ in radians. To reduce the number of response variables from seven to six, we convert the orientation quaternion ($w, p, q, r$) to Euler angles (roll $\phi$, pitch $\theta$, yaw $\psi$). After pre-training, we apply CQR to obtain prediction intervals along each one-dimensional variable, which reveals the predictive uncertainty. Fig.~\ref{fig:CQR} depicts the process for this method.

The main goal of CQR is to produce prediction intervals where the probability of true realizations falling within these intervals is guaranteed to be equivalent to or greater than some chosen coverage rate $p$, i.e., \mbox{$\mathbb{P}\{Y \in C(X) \mid X=x\} \geq p$}. This property is known as \textit{marginal coverage}, and is calculated over all training and test set samples. To compute $C(X)$, we use Random Forest-based quantile regression, which is less prone to over-fitting, requires fewer data and computing, and results in lower and upper bound percentiles that capture heteroskedasticity by expanding and contracting based on the underlying uncertainty in the prediction. The quantiles are then conformalized (corrected) by a calibration set that is split a priori from the training set. The following equations represent the three steps of the procedure:
\begin{equation}
\begin{split}
\{{Q_{l},Q_{h}}\} & \leftarrow QR (\{(X_{i}, Y_{i})\}), \ i \in I_{train}
\end{split}
\end{equation}
\begin{equation}
\begin{split}
E_{i} & \leftarrow \max\{{Q_{l}(X_{i})-Y_{i},Y_{i}-Q_{h}(X_{i})}\}, \ i \in I_{cal}
\end{split}
\end{equation}
\begin{subequations}
\begin{equation}
\begin{split}
C_{cal}(X) & = [Q_{l}(X) - Q_{cal}(E,I_{cal}), \\ & Q_{h}(X) + Q_{cal}(E,I_{cal})], \textrm{ where}
\end{split}
\end{equation}
\begin{equation}
\begin{split}
Q_{cal}&(E,I_{cal}) := (1-\alpha)(1+1/|I_{cal}|)-\textrm{th}\\
& \textrm{quantile of } \{E_i: i \in I_{cal}\}
\end{split}
\end{equation}
\end{subequations}
Here, given a certain miscoverage rate $\alpha$, the quantile regression algorithm ($QR$) fits two conditional quantile functions $Q_{l}$ and $Q_{h}$ on the training set $I_{train}$, which contains samples $X$ and labels $Y$. The effectiveness of the initial prediction interval $[Q_{l}(X),Q_{h}(X)]$ in covering $Y_{i}$ is then measured using conformal scores $E_{i}$, which are evaluated on the held out calibration set $I_{cal}$. If $Y_{i}$ is outside the boundaries, $E_{i}$ measures the distance from the nearest boundary. If $Y_{i}$ falls within the desired boundaries, $E_{i}$ measures the larger of the two distances, accounting for both undercoverage and overcoverage. Finally, the calibrated prediction interval $C_{cal}$ is formed using $Q_{cal}(E,I_{cal})$, which is the empirical $(1-\alpha)(1+1/|I_{cal}|)$-th quantile of ${{E_{i}, \ i \in I_{cal}}}$. Using a calibration set ensures that the resulting intervals have the desired miscoverage rate.

\begin{figure*}[!t]
  \centering
  \includegraphics[width=0.9\linewidth]{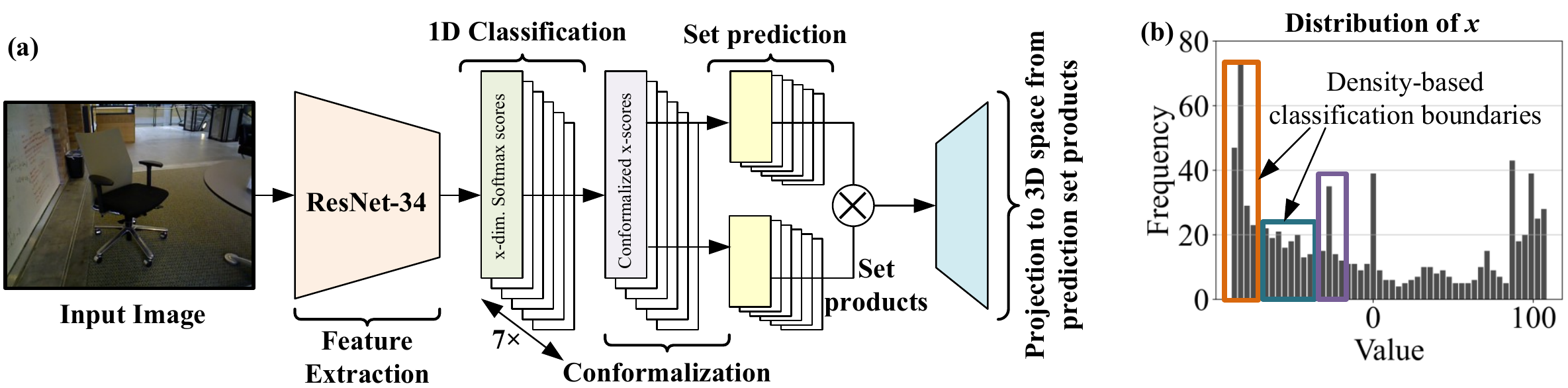}
  \caption{\textbf{Uncertainty-aware VO by conformal set prediction (CSP):} \textbf{(a)} The architecture of CSP for VO. The output of the feature extraction unit is fed to parallel 1D softmax classifiers. Subsequently, classifier outputs are conformalized for the predictive set generation along each dimension. Next, the sets are multiplied and projected to 3D space to generate uncertainty-aware rectangular predictive regions. \textbf{(b)} Our approach also leverages density-aware classification boundaries along each dimension. Based on the training set, frequently visited locations are packed into classes with a narrow gap between upper and lower boundaries for high precision determination when projecting to 3D space. }
  \label{fig:vo_class}
\end{figure*}

\subsection{VO-Uncertainty by Conformalized Set Prediction (CSP)}
This section describes our method, conformalized set prediction (CSP), for converting each of the multiple univariate pose regression problems in VO to pose classification using one-hot encoding and generating conformalized prediction sets by discretizing the flying space. We use the same feature extractor as discussed in the previous section. However, instead of relying on quantile regression to determine the model's uncertainties, we leverage the model's normalized exponential scores (i.e., softmax scores) for conformal set prediction. A notable advantage of this approach is the ability to predict disjoint regions of uncertainty, unlike the previous one that only predicts contiguous regions of uncertainty. 

The approach begins by discretizing the drone's flying space into $K$ sets along each dimension $x$, $y$, and $z$. To achieve this, a non-uniform space discretization is followed, wherein the histograms of the training set trajectories along each dimension are divided into $K$ quantiles to determine class boundaries. The result is a one-hot encoding matrix based on $K$, and a lightweight neural network classifier is built with a programmable output layer of size $K$. For each dimension of the pose, a network is trained to output reliable softmax scores that can be conformalized. Fig.~\ref{fig:vo_class} illustrates the softmax classifiers for each dimension sharing their feature extraction head for parameter sharing and computational efficiency. The product of the predicted set of classes along each dimension produces the net uncertainty regions. Notably, this method can represent frequently visited locations with higher precision, represented by a class with a narrower gap between upper and lower boundaries. Conversely, less frequently visited locations are represented with lower precision.

Under the above classification treatment of VO, to perform conformal set prediction, a calibration set is first separated from the training set to compute conformal scores. It is essential to ensure the statistical guarantee of conformal prediction by ensuring that the average probability of correct classifications within the prediction sets $C(X) \subset {1,...,K}$ is almost exactly $1 - \alpha$, where $\alpha$ denotes an arbitrary miscoverage rate (e.g., 10\%). This marginal coverage property can be expressed mathematically as \mbox{$1 - \alpha \leq \mathbb{P}{Y \in C(X) \mid X=x} \leq 1 - \alpha + \frac{1}{n+1}$}, where $n$ is the size of the calibration set. Conformal scores are then obtained by subtracting the softmax of the correct class for each input from one. Finally, $\hat{q}$ is computed as the $ceil(\frac{(n+1)(1-\alpha)}{n})$ empirical quantile of these conformal scores, and the conformalized prediction set $C(X) \mid X=x$ is formed by incorporating classes with softmax scores greater than $(1-\hat{q})$.

\textit{Importantly}, since the predicted set along each dimension may include classes that are not proximal, the above method has the ability to produce disjoint uncertainty regions. In contrast, the previous method in Sec. III-A (and multivariate methods presented in the later discussion) can only output contiguous uncertainty regions. The class labels in the above conformal set prediction can also be determined by discretizing the entire 3D space; however, the necessary classes for matching precision to dimension-wise discretization would grow exponentially. We also found that the above conformalization procedure may sometimes be too strict, especially when the conditional softmax outputs are not representative. However, it can be improved by using the softmax outputs of all classes in gathering the conformal scores as in \cite{angelopoulos2021}. 

\section{Extracting Multivariate Bands of Predictive Uncertainties in Visual Odometry (VO)}
This section presents two more approaches for extracting predictive uncertainty bands in VO: multivariate conformalized quantile regression with MC-dropout, and conformalized joint prediction. The first approach demonstrates the novel usage of MC-dropout as a data augmentation technique in improving the performance of a conditional variational autoencoder (CVAE). The second approach jointly trains predictive uncertainty bands and pose estimation using a new loss function, thus reducing computational resources for applications where uncertainty estimates are always required. 

\subsection{VO-Uncertainty by Multivariate CQR (MCQR)}  
To generate more informative VO-uncertainty bands that consider the correlation among pose dimensions, we adopt the multivariate conformalized quantile regression algorithm in \cite{feldman2021} and enhance it with MC-dropout (MCQR w/ MC-dropout). This method entails utilizing a CVAE to acquire a proper latent representation of the sample distribution $Y\mid X$ and applying an extension of directional quantile regression (DQR) to create quantile regions. The architecture of the CVAE, shown in Fig.~\ref{fig:customvae}(a), is a variant of a variational autoencoder (VAE) that conditions the generative model on labels during both the encoding and decoding steps. This additional information improves the model's ability to produce targeted outputs. Instead of directly fitting quantile functions to the data with simple regression, a separate neural network is developed to learn the best conditions for sampling the latent representation $Z\mid X$ of the complete multivariate response variable based on $Y\mid X$. At the same time, a quantile region is generated in this dimension-reduced latent space, which has a more informative, flexible, and arbitrary shape when propagated through the decoder.

Compared to the procedure for extracting univariate bands in Sec. III-A, extracting multivariate bands of uncertainty here using a VAE predictor requires more training data due to the model's complexity and higher number of parameters. Therefore, to improve the accuracy of VAE-based uncertainty bands, we utilize MC-dropout to generate additional training data. Training VAEs with the additional data enables uncertainty distillation from the MC-dropout procedure into lightweight conformalized quantile bands. However, the underlying pose estimation model's performance heavily influences this approach's effectiveness. Poor predictions of mean estimates can hinder learning with additional data. 

\subsection{VO-Uncertainty by Conformalized Joint Prediction (CJP)}
This section discusses the joint training of multivariate VO-uncertainty bands and the mean pose, i.e., conformalized joint prediction (CJP). The approach reduces the computational costs of uncertainty-aware VO by sharing parameters and network layers for mean prediction and uncertainty bands, which is particularly useful for platforms where uncertainty estimation must always be ON.

\begin{figure}[!t]
  \centering
  \includegraphics[width=\linewidth]{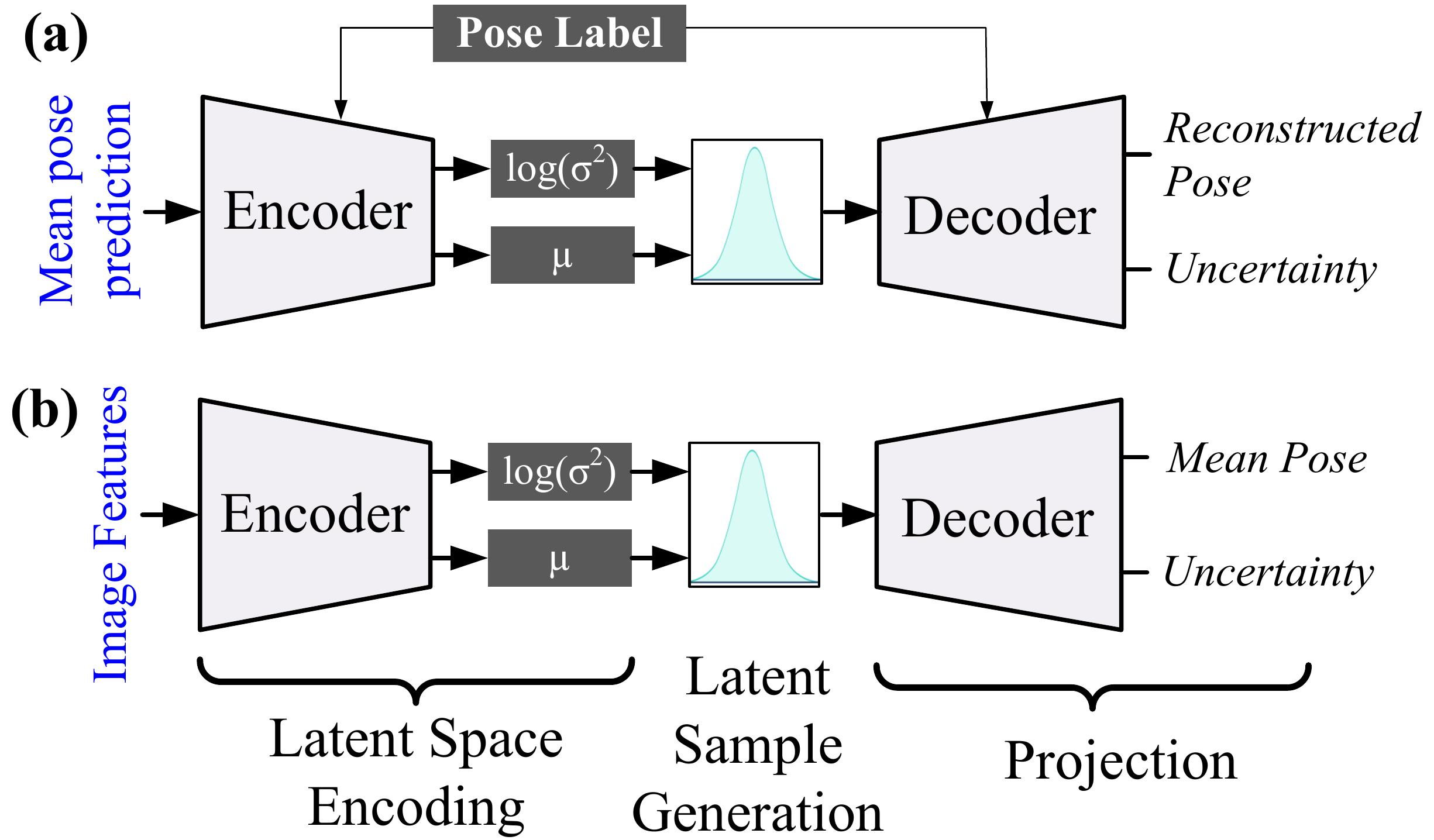}
  \caption{\textbf{VAE architecture for joint training of point (mean) prediction and uncertainty bounds:} \textbf{(a)} Predictions from a pose estimation network (such as PoseNet) are fed to a conditional VAE for pose reconstruction and uncertainty estimation by first encoding to the latent space and then generatively passing through a decoder. \textbf{(b)} Comparatively, the network here jointly learns point prediction and uncertainty estimates directly from the features extracted from camera images. The novel loss function in Eq. (6) is used for the integrated training. The simultaneous extraction of mean and uncertainty bounds provides better computational efficiency for applications where uncertainty awareness must always be ON. }
  \label{fig:customvae}
\end{figure}

Fig.~\ref{fig:customvae}(b) shows the proposed VAE architecture for simultaneously extracting pose predictions and upper/lower uncertainty bounds based on true conditional quantiles. In the network, we adopt a parametric rectified linear unit (PReLU) as an activation function to introduce additional learnable parameters, better handle negative values and avoid the dying ReLU problem (values become zero for any input). In Fig.~\ref{fig:customvae}, the network first extracts features from the input image using a pre-trained convolutional neural network (e.g., MobileNet \cite{howard2017}), followed by encoding to a 3D latent space with a spherical Gaussian probability distribution. Next, a sample is taken from the latent space and propagated through the decoder to generate the multivariate pose and uncertainty-bound estimations at the desired coverage rate.

The pinball loss function, a.k.a. quantile loss, used for conformalized quantile regression in \cite{romano2019,feldman2021} is inadequate for such integrated training of the mean prediction and uncertainty bands. Thus, we propose a new training loss function combining the traditional mean-squared error and Kullback-Leibler (KL) divergence losses of a VAE, $\text{MSE}_{loss}$ and $\text{KL}_{loss}$, with two additional loss components $\text{INTSCORE}_{loss}$ and $\text{COMCAL}_{loss}$. This total loss function, $\mathcal{L}_{Total}$, capitalizes on two unique findings reported in \cite{chung2021}. 

The first finding in the proposed loss function is the unconventional use of the interval score function (i.e., Winkler score \cite{gneiting2007,winkler1972}) to train an uncertainty quantification model to consistently output centered prediction intervals within a specified quantile range. Notably, the interval score function is typically used to evaluate prediction intervals, not optimize them. This approach is advantageous for generating informative prediction intervals over pose trajectories. We choose two percentiles (e.g., $\alpha_{l} =$ 5\% and $\alpha_{h} =$ 95\%) as uncertainty bounds, compute the interval score loss shown below for each percentile, and then take their expected value:
\begin{equation}
\begin{split}
\text{INTSCORE}_{loss} & =(Q_{h}-Q_{l}) + \frac{2}{\alpha}(Q_{l}-y)\mathbb{I}\{y < Q_{l}\} \\
     & + \frac{2}{\alpha}(y-Q_{h})\mathbb{I}\{y > Q_{h}\} \\
\end{split}
\end{equation}
Here, $y$ represents the pose label, $Q_{l}$ and $Q_{h}$ are each dimension's lower and upper quantile estimates, and $\mathbb{I}$ is the indicator function. Including the interval score in $\mathcal{L}_{Total}$ is instrumental in linking pose reconstruction optimization via $\text{MSE}_{loss}$ to quantile optimization over each dimension while honoring the dimensions' covariance and correlation.

The second finding is combined calibration loss, denoted as $\text{COMCAL}_{loss}$. $\text{COMCAL}_{loss}$ establishes an explicit and controllable balance between prediction interval calibration and sharpness rather than relying on the loosely implicit balance in basic pinball loss. $\text{COMCAL}_{loss}$ comprises two objectives: one that minimizes the difference between $p^{cov}_{avg}$ and the chosen marginal coverage rate $p$ ($\text{CAL}_{obj}$), and another that minimizes the distance between $Q_{l}$ and $Q_{h}$ ($\text{SHARP}_{obj}$). $p^{cov}_{avg}$ is the estimated average probability that the pose label values lie within $[\mathit{Q_{l},Q_{h}}]$. The two objectives may conflict with each other since one aims to minimize over-coverage and under-coverage, while the other aims to minimize the length of the prediction interval. This necessitates including a hyper-parameter $\lambda$ to strike the appropriate balance. Thus, $\text{COMCAL}_{loss}$ is defined as follows:
\begin{subequations}
\begin{equation}
\begin{split}
& \text{CAL}_{obj} = \\ 
& \mathbb{I}\{p^{cov}_{avg} < p\} \times \frac{1}{N}\sum_{i=1}^{N}[(y_{i}-Q_{l,h}(x_{i}))\mathbb{I}\{y_{i} > Q_{l,h}\}] + \\ 
& \mathbb{I}\{p^{cov}_{avg} > p\} \times \frac{1}{N}\sum_{i=1}^{N}[(Q_{l,h}(x_{i})-y_{i})\mathbb{I}\{y_{i} < Q_{l,h}\}] 
\end{split}
\end{equation}
\begin{equation}
\begin{split}
\text{SHARP}_{obj} & = \frac{1}{N}\sum_{i=1}^{N}
                \begin{cases} 
                Q_{l}(x_{i}) - Q_{h}(x_{i}),  & \text{$p \leq 0.5$}\\
                Q_{h}(x_{i}) - Q_{l}(x_{i}), & \text{$p > 0.5$}
                \end{cases}
\end{split}
\end{equation}
\begin{equation}
\text{COMCAL}_{loss} = (1-\lambda)\times\text{CAL}_{obj}+\lambda\times\text{SHARP}_{obj}
\end{equation}
\end{subequations}
where $x_{i}$ and $y_{i}$ are individual images and labels of a training batch and $Q_{l,h}$ is meant to denote that $\text{CAL}_{obj}$ is computed separately for both $Q_{l}$ and $Q_{h}$ and then summed.

The complete loss function combining MSE loss, KL divergence, interval score, and calibration loss in our approach is thus given as:
\begin{equation} \label{eq1}
\begin{split}
\mathcal{L}_{Total} & = \text{MSE}_{loss}(y,\hat{y}) + \text{KL}_{loss}(\mu,log(\sigma^2)) \\
& + \text{INTSCORE}_{loss}(y,\mathit{Q_{l},Q_{h}},\{\mathit{\alpha_{l},\alpha_{h}}\}) \\
& + \text{COMCAL}_{loss}(y,p^{cov}_{avg},\mathit{Q_{l},Q_{h}}) \\
\end{split}
\end{equation}
where $\hat{y}$ is the reconstructed pose. $\mu$ and $log(\sigma^2)$ are the hyper-parameters of the VAE's latent space. We estimate $p^{cov}_{avg}$ for each pose dimension with every training batch to enable flexible calibration during training.

Traditional CQR involves splitting the training set into two subsets and conformalizing the quantiles afterward (i.e., split conformal prediction). This is done to avoid full conformal prediction, which requires many more calibration steps than just one, sacrificing speed for statistical efficiency. However, cross-conformal prediction provides a viable alternative for conformalization by striking a reasonable balance between computational and statistical efficiency, utilizing multiple calibration steps across the entire training set \cite{vovk2012}. Our approach is similar, representing a unique form of cross-conformal prediction. We assume that $p^{cov}_{avg}$, which is averaged over randomized training batches, is sufficiently close to an average over the entire training set.

As will be demonstrated later in Sec. V, this approach yields superior outcomes compared to the other methods with a small increase in model complexity and size. The results underscore the method's accuracy, adaptability, and flexibility in producing conformalized prediction intervals for multivariate pose data. The method maintains its strength even when swapping the underlying model from a larger, more accurate one, such as ResNet34, to one better suited for lightweight edge applications like MobileNetV2. Furthermore, its advantages are consistent across multiple datasets, implying its robustness to numerous applications.

\begin{figure*}
\centering
    \includegraphics[width=\linewidth]{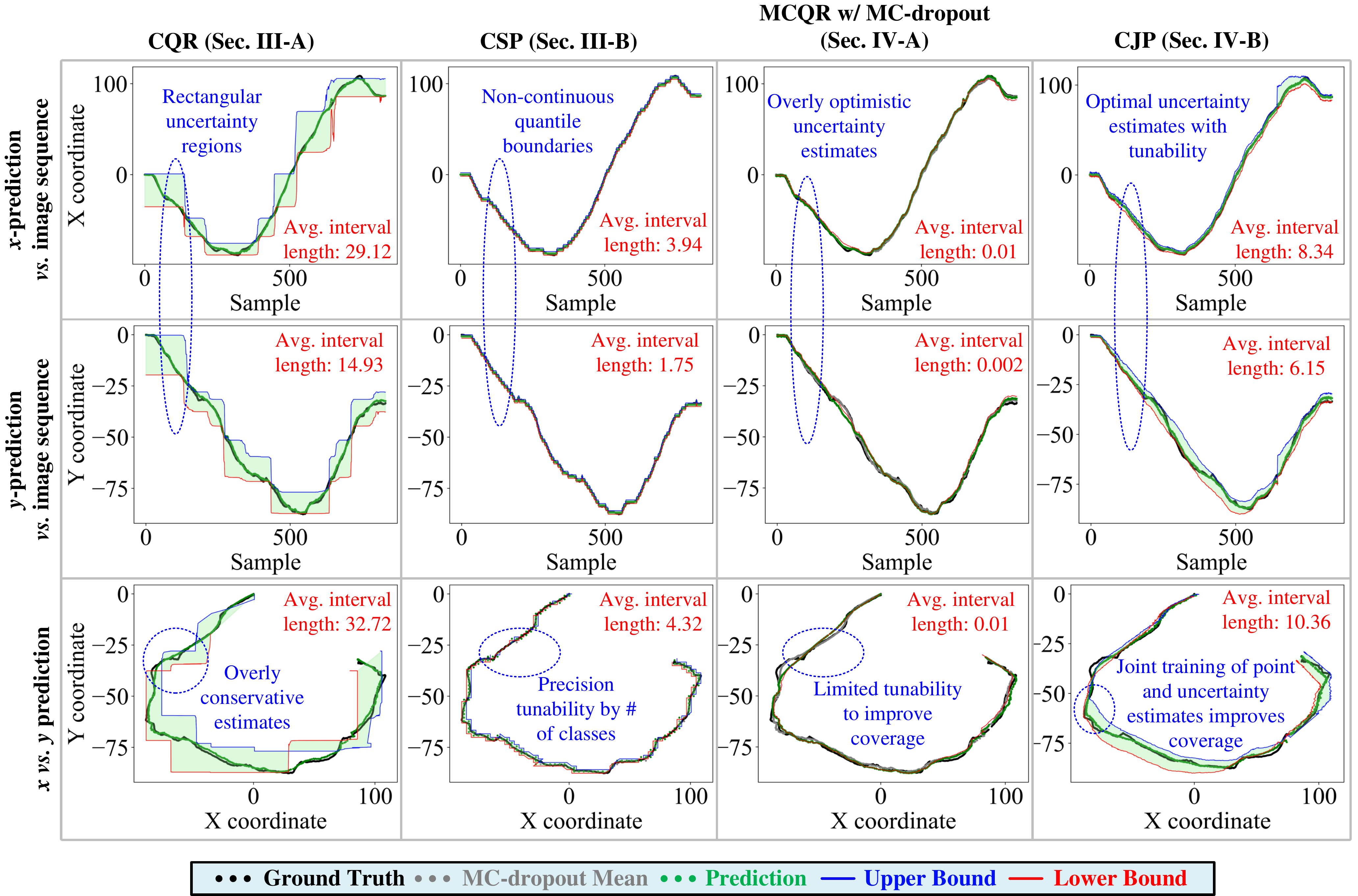}
    \caption{\textbf{Comparison of uncertainty-aware prediction intervals over 1D and 2D pose estimations from the four proposed conformalized visual odometry (CVO) methods in Secs. III and IV:} Differences in average interval length, contiguity, precision, accuracy, coverage, and adaptiveness are highlighted across the methods using Scene 2 of RGB-D Scenes Dataset v.2.}
    \label{fig:all_results}
\end{figure*}

\section{Results and Discussions}
This section presents a chronological review of the results from the four frameworks proposed for conformalized visual odometry (CVO) and compared in Table I. The aim of these methods is to develop reliable prediction intervals that can capture true uncertainty in data-driven pose estimation with guaranteed confidence. While each approach is based on PoseNet \cite{kendall2015}, they can be applied to other pose estimation models. In highlighting the differences between the methods, we use the data from scene 2 of RGB-D Scenes Dataset v.2 \cite{lai2014} as visuals of its pose trajectory are easy to comprehend. All results illustrate uncertainty bands that exhibit at least 90\% guaranteed marginal coverage through CI.

In Sec. III-A, we introduced univariate conformalized quantile regression (CQR) to generate uncertainty bands. However, as shown in Fig.~\ref{fig:all_results}, the resulting prediction interval is overly cautious and box-shaped despite its ability to adjust to the data's heteroscedasticity. This is because the method employs quantile regression on the distinct variables of the multivariate pose response and then merges them without considering their covariance and correlation. As a result, this method lacks sufficient information and is not particularly useful in most cases, despite its low computational complexity. This drawback is consistent across all datasets.

In Sec. III-B, we presented conformalized set prediction (CSP) to generate uncertainty bands. As illustrated in Fig.~\ref{fig:all_results}, this approach adapts well to the data, even with non-continuous uncertainty boundaries. The method's adaptiveness is deeply dependent on the model's softmax scores and the proper discretization of pose values, which can be adjusted by changing the number of classes $K$. Like the first method, this approach has difficulty handling multivariate data, which limits its ability to estimate true uncertainty accurately. Although the average interval length is significantly shorter, the conservativeness of the rectangular uncertainty region has only scaled linearly, limiting its usefulness. 

\begin{figure}
\centering
    \includegraphics[width=\columnwidth]{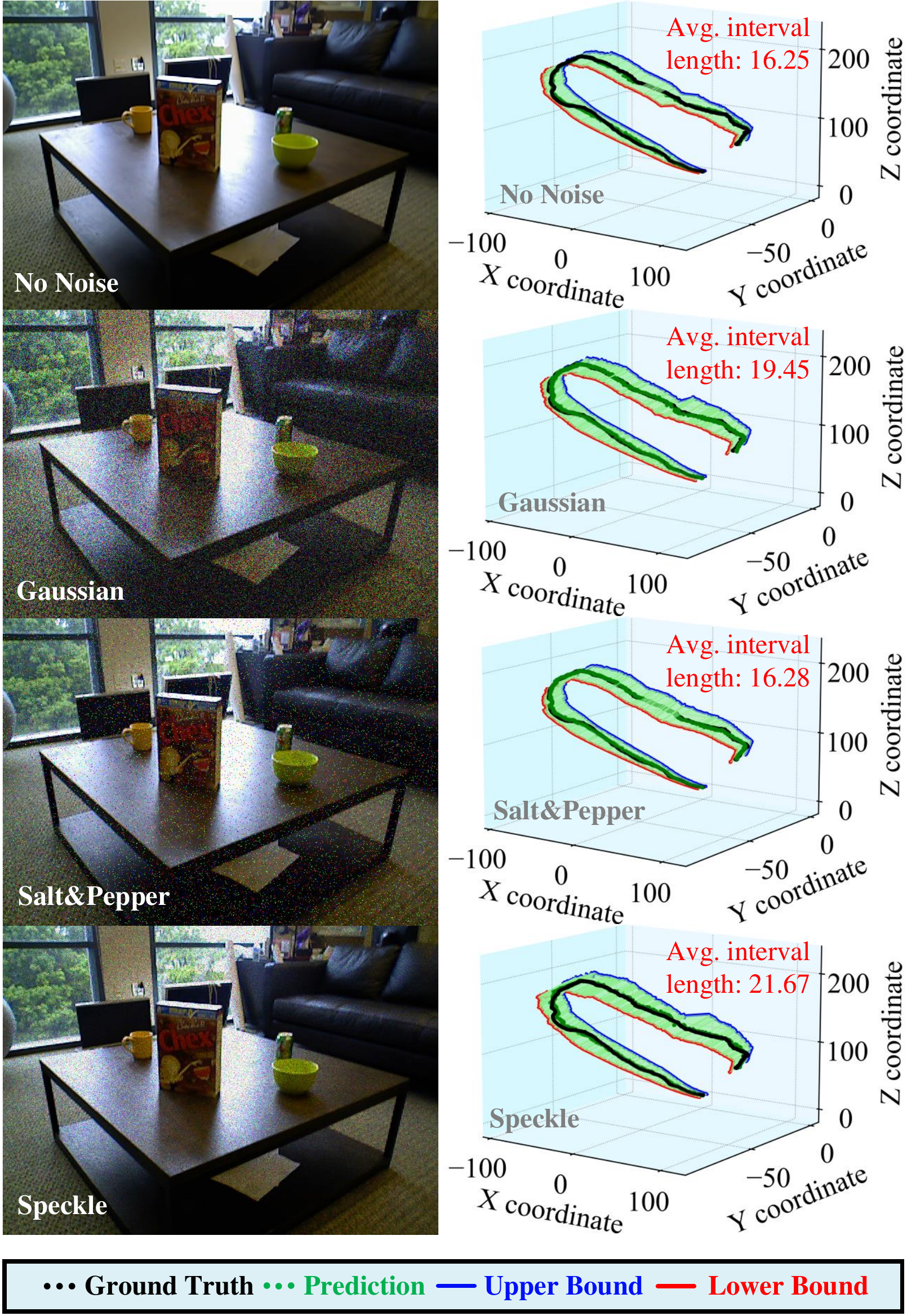}
    \caption{\textbf{Comparison of uncertainty-aware prediction intervals over 3D pose estimations from CJP under different noise conditions:} After switching from the ResNet34 feature extractor to MobileNetV2, gaussian, salt and pepper, and speckle noise were applied to Scene 2 of RGB-D Scenes Dataset v.2 to further stress test the method and demonstrate its resilience. These conditions are meant to simulate real-world disturbance.}
    \label{fig:3dnoise}
\end{figure}

In Sec. IV-A, we introduced the third method, which utilizes multivariate conformalized quantile regression with MC-dropout (MCQR w/ MC-dropout) to generate uncertainty bands. This approach showcases the novel use of MC-dropout as a data augmentation technique, in addition to regularization and uncertainty estimation, to enhance learning. This method is the first step towards addressing the limitations of the previous methods by considering the relationships between pose dimensions. As shown in Fig.~\ref{fig:all_results}, this method has the tightest uncertainty estimates, but they represent the opposite extreme of the conservatism observed in the univariate methods. However, this is inconsistent across all datasets; the method performs much better in some other datasets, indicating that it is highly data-dependent. This inconsistency in uncertainty accuracy is particularly noticeable in smaller datasets, as is the case here. Furthermore, this method has the highest computational complexity, requiring training and optimizing multiple networks. Although it can potentially produce flexible and highly accurate uncertainty intervals, its inconsistency across datasets and computational inefficiency limit its wide applicability.

The final method, described in Sec. IV-B demonstrates a novel combination of multivariate conformalized quantile regression and the mean prediction to produce uncertainty bounds using a new loss function that remarkably achieves multiple objectives simultaneously. We deem this approach conformalized joint prediction (CJP). Fig.~\ref{fig:all_results} demonstrates its significant improvements over the other methods. Firstly, the prediction interval optimally expands and contracts in response to the data, being neither too dispersed nor sharp in each plot. The increase in interval length (and hence uncertainty) at the middle and end points in the trajectory corresponds to worse lighting conditions in the scene's image sequence. Secondly, with the tunable calibration parameter $\lambda$ $\in$ $\{0,1\}$, we can optimize the network for sharpness and/or coverage. Here, $\lambda$ was set to 0.5 to showcase the balanced standard. Additionally, Fig.~\ref{fig:3dnoise} shows results of this method when switching the feature extraction model from ResNet34 to MobileNetV2, effectively reducing its learning capacity by over 80\%. It handles the same image sequence with comparable accuracy even when introducing different forms of noise, such as gaussian, salt and pepper, and speckle. Notably, the uncertainty increases with this transition and even more with added noise. The speckle noise exasperates light variation the most and thus has the largest average interval length. An increase in the average uncertainty intervals indicates that the framework can capture the sensor's uncertainties along with the model's predictive uncertainties. The optimality of this method in extracting true pose estimation uncertainty is consistent across several datasets and conditions, indicating that it can fit many applications.

\section{Conclusion}
This paper introduced and compared four novel frameworks for capturing aleatoric uncertainty in data-driven VO through conformal inference while considering computational resource limitations. The presented framework trade-off essential metrics of uncertainty-aware inference, such as computational workload, adaptiveness of uncertainty interval, training time, \textit{etc.}, aiming to provide a spectrum of solutions against widely varying robustness and resource constraints on edge robotics. We discussed the architectures of generative deep neural networks for estimating multivariate uncertainty bands along with point (mean) prediction, data augmentation techniques to improve the accuracy of uncertainty estimation, and novel training loss functions that combined interval scoring and calibration loss with traditional training metrics such as KL-divergence to improve uncertainty-aware learning. 

\bibliographystyle{IEEEtran}
\bibliography{references}

\end{document}